\documentclass[10pt, a4paper]{article}
\usepackage{amsmath} 
\usepackage{amssymb}
\usepackage{booktabs}
\usepackage{tabularx}
\usepackage{arydshln}
\usepackage[table]{xcolor}
\usepackage{multirow} 

\usepackage[final]{lrec2026}

\makeatletter
\def\ps@plain{\ps@empty} 
\makeatother
\title{CRiT-QA: Evaluating Multi-hop Reasoning with \\ Counterfactual Chains and Distractor Traps}

\name{JungMin Yun\textsuperscript{\rm 1}\textsuperscript{\rm *}, JuneHyoung Kwon\textsuperscript{\rm 1}\textsuperscript{\rm *}\thanks{*Equal contribution.}, YoungBin Kim\textsuperscript{\rm 1, 2}} 
\address{
    \textsuperscript{\rm 1} Department of Artificial Intelligence, Chung-Ang University\\
    \textsuperscript{\rm 2} Graduate School of Advanced Imaging Sciences, Multimedia and Film, Chung-Ang University\\
    \{cocoro357, dirchdmltnv, ybkim85\}@cau.ac.kr\\
}

\abstract{
Evaluating the multi-hop reasoning capabilities of large language models remains a significant challenge. Although current models achieve strong results on existing multi-hop question answering datasets, such performance often masks two critical vulnerabilities: (1) reliance on internal parametric knowledge rather than adherence to the provided context, and (2) exploitation of dataset shortcuts, such as single-document cues or type-matching, that diminish the need for genuine evidence aggregation across multiple documents. We introduce CRiT-QA (Counterfactual Reasoning with Traps), a dataset explicitly designed to address both limitations. To neutralize reliance on memorized knowledge and enforce strict context dependency, CRiT-QA transforms factual reasoning chains with counterfactual entities. Furthermore, it injects multi-anchor distractor chains, plausible but incorrect reasoning paths that diverge at different hops. These traps require models to follow the entire reasoning process rather than exploiting shallow heuristics. Our experiments show that LLMs exhibit substantial performance degradation on CRiT-QA compared to standard datasets, exposing their vulnerability to counterfactual conditions and distractor traps. CRiT-QA thus serves as a rigorous diagnostic tool for evaluating genuine multi-hop reasoning and provides a foundation for developing more reliable, evidence-grounded
LLMs.
 \\ \newline  }

\begin{document}

\maketitleabstract

\section{Introduction}

Retrieval-Augmented Generation (RAG) has become a dominant approach for enhancing large language models (LLMs) by grounding outputs in external knowledge sources rather than relying solely on internal parametric memory~\cite{lewis2020retrieval, fan2024survey}. This paradigm has demonstrated promising improvements in factuality and adaptability across diverse domains~\cite{tang2024multihop, gao2024retrievalaugmentedgenerationlargelanguage}. The effectiveness of RAG systems, however, critically depends on the model's ability to identify, link, and aggregate multiple pieces of evidence distributed across different documents~\cite{fang-etal-2024-trace, suryawanshi2025knowledge}. 

Multi-hop Question Answering (QA) has thus become a standard task for evaluating the higher-order reasoning capabilities of LLMs. In principle, such datasets are designed to assess whether models can identify and synthesize intermediate evidence to derive logically coherent final answers~\cite{kim-etal-2024-improving-multi, li2024meqa, liu2025hopragmultihopreasoninglogicaware}. However, recent studies have revealed that the strong performance of LLMs on existing multi-hop QA datasets does not necessarily reflect genuine reasoning ability~\cite{wu2024mrke, jiang-etal-2024-peek, parmar-etal-2024-logicbench}. Instead, models often exploit artifacts in the dataset or rely on shallow surface-level patterns, thereby inflating their performance without demonstrating robust reasoning~\cite{schlegel-etal-2020-framework, trivedi-etal-2022-musique, suryawanshi2025knowledge}.

\begin{table}[t]
\centering
\renewcommand{\arraystretch}{1.4}
\small
\begin{tabularx}{\columnwidth}{@{} c X @{}}
\toprule

\textbf{(A)} & \textbf{Question:} Who founded the company that distributed the film UHF? \newline
      \textcolor{gray}{\footnotesize\textit{------ without any context ------}} 
      \vspace{0.2em} \newline
      \textbf{LLM-generated Answer:} Mike Medavoy \textcolor{green!60!black}{\textbf{\checkmark}} \\
\midrule
\textbf{(B)} & \textbf{Question:} In which \textcolor{orange!80!black}{\textbf{county}} is Mark Dismore's birthplace located? \newline
      \textbf{Paragraph 1:} Greenfield is a city in and the county seat of \textcolor{orange!80!black}{\textbf{Hancock County}}, Indiana, United States, and a part of the Indianapolis metropolitan area. (...)
      \vspace{0.2em} \newline
      \textbf{LLM-generated Answer:} Hancock County \textcolor{green!60!black}{\textbf{\checkmark}}
      \vspace{0.5em} \newline
      \textcolor{gray}{\textit{\textbf{Sub-question:} What is the place of birth of Mark Dismore?}} \newline
      \textcolor{gray}{\textit{\textbf{LLM-generated Answer:}}} \textcolor{red!80!black}{\textbf{\textit{Unanswerable}}} \\
\bottomrule
\end{tabularx}
\caption{Examples illustrating evaluation vulnerabilities in multi-hop QA: (A) an LLM answering correctly without context, and (B) a model exploiting single-paragraph cues to bypass the full multi-hop reasoning process.}
\label{tab:examples}
\end{table}

From the model perspective, LLMs often bypass the provided context and instead rely on internally memorized parametric knowledge acquired during pretraining~\cite{wucofca, bi2025parameters, cheng2024understanding}. This tendency undermines the validity of the evaluation, as models may generate correct answers without engaging with the evidence. For example, as illustrated in Table~\ref{tab:examples}(A), an LLM can answer "Who founded the company that distributed the film \textit{UHF}?" correctly as "Mike Medavoy," even when no context is supplied, indicating reliance on internal memory rather than reasoning over the provided passages. Moreover, large-scale pretraining poses risks of data leakage, leading to contamination that further compromises the reliability of reasoning evaluation.

On the dataset side, existing benchmarks contain shortcuts that enable models to arrive at correct answers without completing the full reasoning process~\cite{guo-etal-2023-counterfactual, yang-etal-2024-large-language-models, ho-etal-2023-analyzing}. A question phrased with a type cue such as "which person" can sometimes be solved by selecting the only person entity in a passage; in other cases, a single paragraph suffices, and cross-document aggregation is unnecessary. Table~\ref{tab:examples}(B) demonstrates this failure case for a question about "Mark Dismore's birthplace," where the model produces "Hancock County" by matching the "county" type in the question to a surface string in the first paragraph, while failing to resolve the requisite intermediate sub-question about the actual place of birth. Such artifacts encourage heuristic pattern matching and obscure whether models truly perform multi-hop reasoning.

To address these limitations, we introduce CRiT-QA (Counterfactual Reasoning with Traps), a dataset designed to rigorously evaluate LLM reasoning under counterfactual and distractor conditions. CRiT-QA targets both weaknesses in a systematic manner. First, to reduce dependence on internal knowledge, factual reasoning chains are transformed using counterfactual entities so that correct answers cannot be recalled from pretraining memory and must be inferred strictly from the given context. Second, to overcome dataset shortcuts, CRiT-QA injects multi-anchor distractor traps, plausible but ultimately incorrect reasoning paths that diverge at different hops while maintaining type consistency via named entity recognition. These distractors prevent models from relying on shallow cues and require stepwise verification of the evidence path.

Our experiments reveal that existing LLMs experience substantial performance degradation on CRiT-QA, standing in sharp contrast to their strong results on traditional datasets. This drop is consistent across both proprietary and open-source models and becomes more severe as the required reasoning chain length increases. For example, even the strongest proprietary systems, such as \texttt{GPT-4o} and \texttt{Gemini-2.5-Pro}, show marked declines when moving from 2-hop to 4-hop questions. The performance collapse is even more dramatic for open-source models like \texttt{LLaMA-3-8B} and \texttt{Qwen2.5-7B}, whose accuracy is reduced by more than half. These findings indicate that CRiT-QA successfully stresses models beyond surface-level recall, exposing their vulnerability to counterfactual reasoning constraints and the increasing density of distractor traps as reasoning depth grows. In summary, our main contributions are as follows:

\begin{enumerate}
    \item We propose CRiT-QA, a diagnostic multi-hop QA dataset that simultaneously addresses two major limitations of existing evaluations: overreliance on internal parametric knowledge and the exploitation of dataset shortcuts.
    \item We design an automated data construction pipeline leveraging LLMs to transform existing datasets into counterfactual versions and to generate multi-anchor distractor chains, thereby creating more challenging contexts for evaluating reasoning.
    \item We empirically demonstrate that state-of-the-art LLMs, despite excelling on standard datasets, substantially underperform on CRiT-QA, underscoring the persistent gap between surface-level success and genuine multi-hop reasoning ability.
\end{enumerate}

Ultimately, CRiT-QA serves as a rigorous diagnostic tool for evaluating genuine multi-hop reasoning in LLMs. By exposing critical vulnerabilities to counterfactuals and distractors, it provides a robust foundation for developing more reliable, evidence-grounded language models.

\begin{figure*}[t!]
  \centering
  \includegraphics[width=1.0\textwidth]{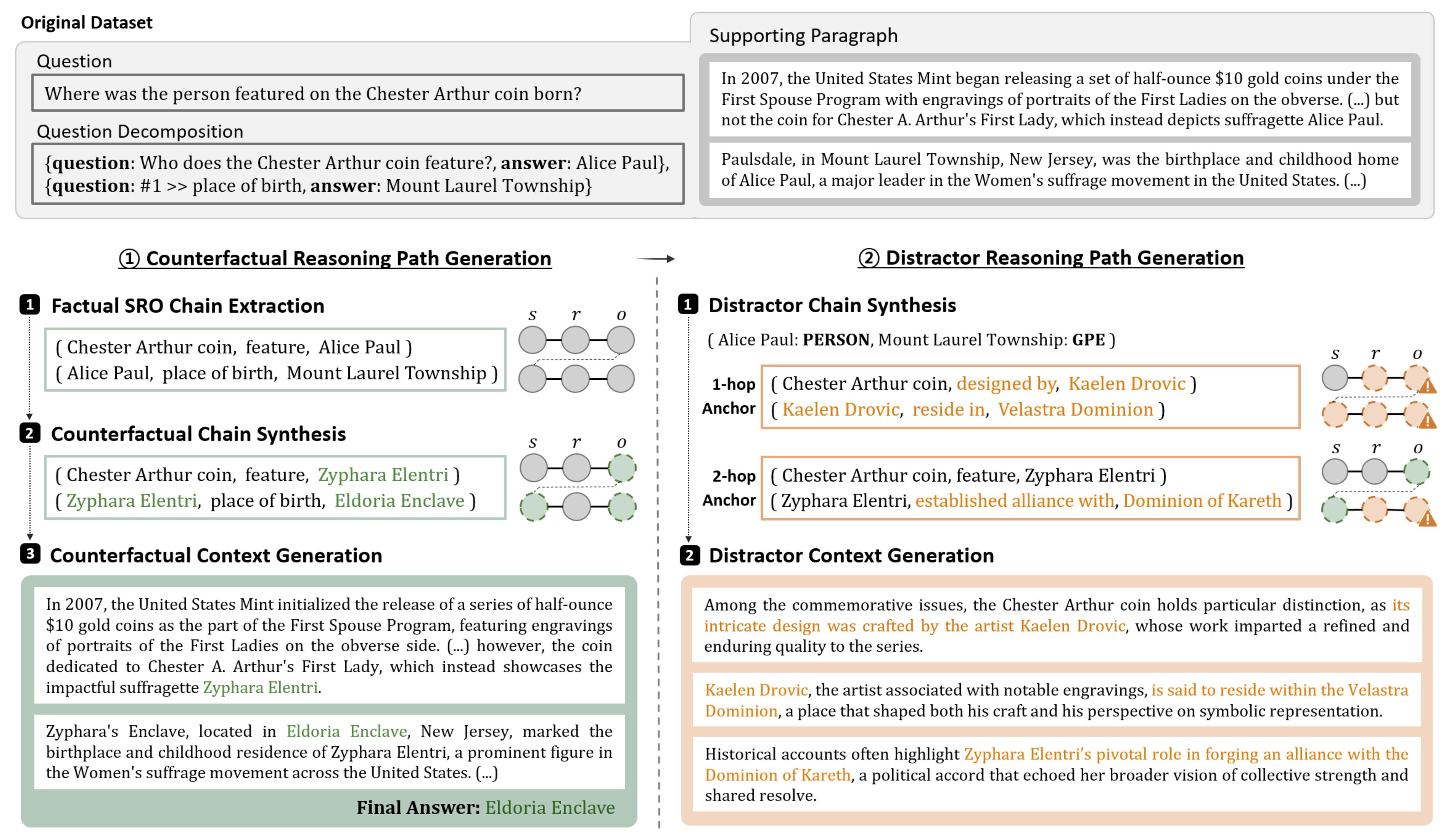}
  \caption{Overall pipeline of the proposed CRiT-QA dataset construction.}
\end{figure*}

\section{Related Work}
\label{sec:append-how-prod}

\paragraph{Multi-hop QA Datasets.} Multi-hop QA datasets have been widely adopted as a standard benchmark for evaluating a model's ability to perform complex reasoning by integrating evidence from multiple sources~\cite{press-etal-2023-measuring, geva-etal-2021-aristotle, tang2024multihop}. HotpotQA~\cite{yang-etal-2018-hotpotqa} requires multi-hop reasoning across linked Wikipedia articles, with annotated supporting facts that enable explainable evaluation. 2WikiMultiHopQA~\cite{ho-etal-2020-constructing} systematically pairs two Wikipedia pages to focus on structured cross-document reasoning, while MuSiQue~\cite{trivedi-etal-2022-musique} constructs multi-hop questions from single-hop components, providing explicit decompositions into sub-questions with intermediate answers. 

\paragraph{Reasoning under Distractors.} Although these datasets have substantially advanced multi-hop QA evaluation, subsequent analyzes have revealed that models often exploit dataset-specific artifacts, such as type matching or single-paragraph cues, or rely on memorized parametric knowledge~\cite{bhuiya2024seemingly, li2024meqa, schnitzler2024morehopqa}. As a result, models can achieve inflated performance without genuine multi-hop reasoning. For instance, many questions can be solved through simple pattern matching or even one-hop shortcuts, as question terms frequently overlap with tokens in the answer sentences~\cite{schlegel-etal-2020-framework, bhuiya2024seemingly}. 

To mitigate these weaknesses, several datasets have been designed to deliberately introduce distractions into the reasoning process. One of the works~\cite{jiang-bansal-2019-avoiding} proposes adversarial data construction strategies that compel models to follow the correct reasoning chain, while other works~\cite{ding2021reasoning} investigate reasoning robustness under fact-level conflicts. ClashEval~\cite{wu2024clasheval}, for example, perturbs retrieved passages with subtle or overt errors to quantify the interplay between a model's internal knowledge and external evidence. Another study~\cite{bhuiya2024seemingly} augments multi-hop datasets by injecting semantically similar distractor paragraphs, creating plausible yet misleading reasoning paths to evaluate robustness.

\paragraph{Reasoning under Counterfactual Knowledge.}
Beyond distractor-based evaluation, a complementary line of research investigates reasoning under counterfactual conditions. This paradigm probes whether models can prioritize contextual reasoning over potentially conflicting parametric knowledge~\cite{frohberg-binder-2022-crass}. By introducing hypothetical alternatives to established facts, counterfactual scenarios challenge models to perform genuine reasoning rather than mere factual recall~\cite{chen2025counterbenchbenchmarkcounterfactualsreasoning}. DisentQA~\cite{neeman-etal-2023-disentqa} adopts counterfactual data augmentation to disentangle parametric and contextual knowledge, although it remains limited to single-hop settings. Other works manipulate the question itself: CREPE~\cite{yu-etal-2023-crepe} incorporates false presuppositions into questions to evaluate whether models can detect and correct them, and IFQA~\cite{yu-etal-2023-ifqa} embeds counterfactual presuppositions directly in questions, requiring models to integrate these assumptions with retrieved factual evidence in an open-domain setting. While these approaches provide valuable diagnostic probes, they are confined to local, question-level edits rather than assessing reasoning consistency across an entire multi-hop reasoning chain. Closest to our work, CofCA~\cite{wucofca} leverages an LLM to rewrite original documents with counterfactual content and subsequently generates new multi-hop QA pairs from the modified context. However, CofCA focuses on regenerating new questions aligned with rewritten documents, while we deliberately retain the original multi-hop questions and systematically replace the entire underlying reasoning chain along with its supporting evidence, ensuring that the original compositional reasoning demands remain intact. Moreover, we combine counterfactual chain replacement with multi-anchor distractors within a single unified framework, enabling a controlled, incremental escalation of reasoning difficulty that neither challenge alone can provide.

\section{Dataset Construction Pipeline}

We construct CRiT-QA through a multi-stage, LLM-driven pipeline designed to rigorously evaluate reasoning in multi-hop QA under both counterfactual and distracting conditions. Within this pipeline, factual reasoning chains are transformed into counterfactual and distractor variants, ensuring that models engage in authentic reasoning processes while preventing reliance on shallow inference driven by memorized associations or surface-level patterns.
 
\subsection{Counterfactual Reasoning Path}
\paragraph{Factual SRO Chain Extraction.} We begin with the MusiQue dataset~\cite{trivedi-etal-2022-musique}, which is suitable for our setting, as it explicitly provides step-by-step decompositions for multi-hop questions. Each sample consists of a multi-hop question $Q$, a final factual answer $A_{fact}$, a set of supporting paragraphs $P$, and a reasoning decomposition $D$. The decomposition consists of an ordered sequence of sub-questions $q_i$ and their corresponding sub-answers $a_i$. We first filter the paragraphs to retain only the gold evidence paragraphs $P_{fact}$ that support this decomposition. We then employ an LLM to transform the decomposition $D$ into a structured Subject--Relation--Object (SRO) chain $C_{fact}$. This chain is a sequence of $N$ hops, $C_{fact}=\{h_1, h_2, \ldots, h_N\}$, where each hop $h_i$ is a triple $(s_i, r_i, o_i)$. The relation $r_i$ is derived from the sub-question $q_i$, and the object $o_i$ corresponds to the sub-answer $a_i$. The subject $s_i$ is either an entity from the original question $Q$ or the object $o_k$ ($k<i$) of a previous hop, explicitly linking the reasoning steps. This LLM-guided mapping enables us to construct an explicit and coherent reasoning chain $C_{fact}$ that faithfully captures the stepwise, and potentially non-linear, reasoning trajectory.

\begin{table*}[t!]
\centering
\resizebox{1.0\textwidth}{!}{%
\begin{tabular}{l|cc cc|cc cc}
\toprule
& \multicolumn{2}{c}{\textit{CRiT-QA (Ours)}}
& \multicolumn{2}{c|}{\textit{MuSiQue}}
& \multicolumn{2}{c}{\textit{2WikiMultiHopQA}}
& \multicolumn{2}{c}{\textit{HotpotQA}} \\
\cmidrule(lr){2-3} \cmidrule(lr){4-5} \cmidrule(lr){6-7} \cmidrule(lr){8-9}
& \textbf{EM} & \textbf{F1}
& \textbf{EM} & \textbf{F1}
& \textbf{EM} & \textbf{F1}
& \textbf{EM} & \textbf{F1} \\
\midrule\midrule
\texttt{LLaMA-3-8B} & 16.94 & 23.29 & 24.74 & 36.52 & 35.06 & 44.52 & 38.51 & 52.63 \\
\texttt{Longchat-13B-16k} & 11.14 & 17.83 & 18.54 & 31.39 & 26.91 & 38.58 & 30.24 & 44.37 \\
\texttt{Mistral-7B-Instruct} & 18.55 & 31.26 & 27.51 & 46.98 & 35.37 & 50.86 & 44.62 & 63.45 \\
\texttt{Qwen2.5-7B} & 19.30 & 25.20 & 34.96 & 47.36 & 42.76 & 50.65 & 52.02 & 65.54 \\
\midrule
\texttt{GPT-3.5-Turbo} & 30.93 & 40.51 & 39.68 & 55.28 & 50.02 & 62.13 & 57.30 & 72.90 \\
\texttt{GPT-4o} & 39.59 & 48.52 & 47.66 & 63.69 & 68.66 & 80.22 & 63.70 & 79.98 \\
\texttt{Gemini-2.5-Flash} & 38.30 & 46.24 & 55.56 & 69.37 & 71.49 & 79.99 & 66.21 & 81.27 \\
\texttt{Gemini-2.5-Pro} & 44.27 & 53.89 & 59.08 & 73.99 & 73.46 & 82.90 & 66.28 & 81.50 \\
\texttt{Claude Haiku 4.5} & 30.02 & 37.45 & 48.37 & 63.62 & 58.06 & 69.87 & 59.11 & 75.89 \\
\texttt{Claude Sonnet 4.5} & 40.25 & 48.96 & 54.86 & 70.66 & 68.62 & 77.89 & 64.16 & 80.42 \\
\bottomrule
\end{tabular}%
}
\caption{Performance comparison of LLMs across CRiT-QA and three standard multi-hop QA benchmarks.}
\label{tab:benchmark_results}
\end{table*}

\paragraph{Counterfactual Chain Synthesis.} To introduce systematic counterfactual variation, we employ an LLM to transform the factual chain $C_{fact}$ into a corresponding counterfactual chain $C_{cf}=\{h^{(c)}_1, h^{(c)}_2, \ldots, h^{(c)}_N\}$, where each hop is $h^{(c)}_i=(s^{(c)}_i,r^{(c)}_i,o^{(c)}_i$). This synthesis is guided by the following principles to ensure coherence and logical consistency. First, each factual object $o_i$ is replaced by a newly generated fictional counterfactual object $o^{(c)}_i$. Second, subjects originating from the original question $Q$ and all relations are preserved to maintain structural alignment and logical flow. Third, if the subject $s_i$ corresponds to the object $o_k$ of a preceding hop, the new subject $s^{(c)}_i$ is set to the corresponding counterfactual object $o^{(c)}_k$, ensuring the internal consistency of the reasoning chain.

\paragraph{Counterfactual Context Generation.} To provide coherent contextual grounding, we prompt an LLM to rewrite the supporting paragraphs $P_{fact}$ into counterfactual contexts $P_{cf}$. This rewriting process goes beyond simple entity substitution: the model adjusts syntax, discourse flow, and narrative structure to ensure grammatical correctness, stylistic coherence, and logical alignment with the counterfactual reasoning chain $C_{cf}$. The resulting passages $P_{cf}$ seamlessly integrate fictional entities into the narrative, producing a self-consistent and contextually grounded representation. In addition, the LLM generates the counterfactual final answer $A_{cf}$ in alignment with the newly constructed $P_{cf}$ and reasoning chain $C_{cf}$, ensuring consistency across all components of the sample. 

At the conclusion of this stage, each sample is represented by four components: the original question $Q$, the counterfactual reasoning chain $C_{cf}$, the rewritten contexts $P_{cf}$, and the counterfactual final answer $A_{cf}$.

\subsection{Distractor Reasoning Path}
To evaluate model robustness against misleading alternatives, we generate distractor reasoning chains that diverge from the correct counterfactual chain. This stage generates alternative reasoning paths that closely mimic the structural form of valid reasoning chains but deliberately diverge at specific hops, thereby producing coherent yet ultimately incorrect reasoning trajectories.

\paragraph{Distractor Chain Synthesis.} 
We prompt the LLM to generate a set of distractor chains, each diverging from the correct counterfactual chain at a different hop. Given an $N$-hop counterfactual chain $C_{cf}$, we construct $N$ distinct distractor chains $C_{dist}^{(j)}$ (where $j \in [1,N]$), with hop $j$ serving as its initial divergence point. Each distractor hop is denoted as $h_i^{(d)}=(s_i^{(d)}, r_i^{(d)}, o_i^{(d)})$ and is generated in a hop-by-hop manner by replacing counterfactual elements with plausible but misleading alternatives. To ensure type consistency, we adopt an NER-driven strategy: the LLM analyzes the entity type of each object (e.g., person, location, date) and generates a new distractor object of the same type. Relations are modified to be semantically distinct from the originals. The subject is either preserved or replaced with the corresponding distractor object from a preceding hop, thereby maintaining coherent linkage across the reasoning chain. This approach yields distractor paths that are structurally valid and contextually plausible, yet ultimately lead to incorrect reasoning trajectories unrelated to the correct answer.

\paragraph{Distractor Context Generation.} 
To provide textual grounding for the distractor chains, we generate additional paragraphs corresponding to their unique hops. Specifically, we identify the set of all unique distractor hops that appear in any distractor chain but not in the correct counterfactual chain. For each such hop, we prompt an LLM to generate a short, coherent paragraph that provides supporting evidence. The resulting set of distractor paragraphs $P_{dist}$ is then combined with the counterfactual paragraphs $P_{cf}$ to form the final context $P_{final} = P_{cf} \cup P_{dist}$. This integrated context introduces multiple competing facts that are all textually plausible, compelling models to carefully discriminate between correct and misleading reasoning paths without relying on simple surface-level patterns.

\begin{table*}[t]
\centering
\resizebox{0.9\textwidth}{!}{%
\begin{tabular}{l|cc|cc}
\toprule
\multirow{2}{*}{\textbf{Setting}} & \multicolumn{2}{c|}{\texttt{(Qwen2.5-7B)}} & \multicolumn{2}{c}{\texttt{(Gemini-2.5-Pro)}} \\ 
 & \textbf{EM} & \textbf{F1} & \textbf{EM} & \textbf{F1} \\ \midrule \midrule
\textit{Oracle ($Q + P_{fact}$)} & 34.96 & 47.36 & 59.08 & 73.99 \\ \midrule
\textit{with Counterfactuals ($P_{cf}$)} & 24.77 \textcolor{red!80!black}{\scriptsize ($\downarrow$10.19)} & 31.18 \textcolor{red!80!black}{\scriptsize ($\downarrow$16.18)} & 50.86 \textcolor{red!80!black}{\scriptsize ($\downarrow$8.22)} & 62.02 \textcolor{red!80!black}{\scriptsize ($\downarrow$11.97)} \\ 
\quad \quad \textit{with Distractors ($P_{cf}+P_{dist}$)} & 19.30 \textcolor{red!80!black}{\scriptsize ($\downarrow$5.47)} & 25.20 \textcolor{red!80!black}{\scriptsize ($\downarrow$5.98)} & 47.74 \textcolor{red!80!black}{\scriptsize ($\downarrow$3.12)} & 57.33 \textcolor{red!80!black}{\scriptsize ($\downarrow$4.69)} \\ \midrule
\textit{without Context (only $Q$)} & 0.00  & 1.27 & 0.00  & 1.74 \\ \bottomrule
\end{tabular}%
}
\caption{Ablation results on CRiT-QA under different evidence settings.}
\label{tab:ablation}
\end{table*}

\section{Experiments}
\subsection{Experimental Setup}
We evaluate LLMs on our constructed CRiT-QA dataset using a diverse set of models. For open-source models, we include \texttt{LLaMA-3-8B}~\cite{dubey2024llama}, \texttt{Qwen2.5-7B}~\cite{qwen2025qwen25technicalreport}, \texttt{Longchat-13B-16k}\footnote{https://huggingface.co/lmsys/longchat-13b-16k}, and \texttt{Mistral-7B-Instruct}~\cite{jiang2023mistral7b}. For API-based commercial models, we evaluate \texttt{GPT-3.5-Turbo}~\cite{brown2020languagemodelsfewshotlearners}, \texttt{GPT-4o}~\cite{hurst2024gpt}, \texttt{Gemini-2.5-Flash}, \texttt{Gemini-2.5-Pro}~\cite{comanici2025gemini}, \texttt{Claude Haiku 4.5}, and \texttt{Claude Sonnet 4.5}. To assess multi-hop QA performance, we report both Exact Match (EM) and F1 scores. In all experiments, we explicitly instruct the LLMs to derive their final answers based solely on the provided context, thereby reducing the influence of their internal parametric knowledge.

\begin{table*}[t!]
\centering
\resizebox{0.83\textwidth}{!}{%
\begin{tabular}{l|cc|cc|cc}
\toprule
& \multicolumn{2}{c|}{\textbf{\textit{2-hop}}} & \multicolumn{2}{c|}{\textbf{\textit{3-hop}}} & \multicolumn{2}{c}{\textbf{\textit{4-hop}}} \\
\cmidrule(lr){2-3} \cmidrule(lr){4-5} \cmidrule(lr){6-7}
 & \textbf{EM} & \textbf{F1} & \textbf{EM} & \textbf{F1} & \textbf{EM} & \textbf{F1} \\ \midrule \midrule
\texttt{LLaMA-3-8B}          & 20.88 & 28.91 & 15.53 & 21.05 & 7.41  & 10.18 \\
\texttt{Longchat-13B-16k}    & 13.84 & 22.69 & 10.13 & 14.87 & 5.93  & 9.11  \\
\texttt{Mistral-7B-Instruct} & 22.08 & 35.55 & 15.13 & 28.26 & 14.07 & 23.64 \\
\texttt{Qwen2.5-7B}          & 22.80 & 26.69 & 16.97 & 22.49 & 12.84 & 16.44 \\ \midrule
\texttt{GPT-3.5-Turbo}       & 34.80 & 44.82 & 30.13 & 40.18 & 20.49 & 26.24 \\
\texttt{GPT-4o}              & 43.84 & 52.78 & 36.71 & 47.87 & 31.85 & 36.53 \\
\texttt{Gemini-2.5-Flash}    & 40.56 & 49.63 & 38.29 & 46.54 & 31.36 & 35.22 \\
\texttt{Gemini-2.5-Pro}      & 44.89 & 55.56 & 45.13 & 55.03 & 40.74 & 46.48 \\
\texttt{Claude-4.5-Haiku}    & 32.16 & 41.12 & 29.87 & 36.27 & 23.70 & 28.32 \\
\texttt{Claude-4.5-Sonnet}   & 43.52 & 52.82 & 39.74 & 49.47 & 31.11 & 36.07 \\ \bottomrule
\end{tabular}%
}
\caption{Performance comparison on CRiT-QA according to hop length.}
\label{tab:benchmark_results_hop}
\end{table*}

\subsection{Experimental Results}
In Table~\ref{tab:benchmark_results}, we report the performance of various LLMs on CRiT-QA in comparison to three widely used multi-hop QA datasets\footnote{All experiments were conducted on the dev sets of the respective datasets. For MuSiQue, we used the \textit{musique\_ans\_v1.0\_dev}, which contains only answerable QA pairs.}: 2WikiMultiHopQA~\cite{ho-etal-2020-constructing}, MuSiQue~\cite{trivedi-etal-2022-musique}, and HotpotQA~\cite{yang-etal-2018-hotpotqa}. We note that this comparison is not intended as a strict head-to-head evaluation across datasets, given the inherent structural differences across datasets. Rather, its purpose is to illustrate a broader trend: models that achieve strong performance on standard multi-hop benchmarks exhibit substantial performance degradation when faced with the counterfactual and distractor-based challenges introduced by CRiT-QA.

\paragraph{Main Results.} We observe a consistent decline across all evaluated models, including both open-source and API-based commercial systems. This universal performance gap highlights a fundamental vulnerability in current reasoning capabilities. For instance, \texttt{Gemini-2.5-Pro} achieves the highest performance on CRiT-QA (44.27 EM and 53.89 F1); however, this still represents a substantial drop from its scores on MuSiQue (59.08 EM) and 2WikiMultiHopQA (73.46 EM). Similarly, \texttt{GPT-4o} and \texttt{Claude Sonnet 4.5} fall from 47.66 and 54.86 EM on MuSiQue to 39.59 and 40.25 EM on CRiT-QA, respectively. 

Furthermore, this vulnerability is even more pronounced among open-sourced models. For instance, \texttt{Qwen2.5-7B} attains 52.02 EM on HotpotQA and 34.96 EM on MuSiQue but only 19.30 EM on CRiT-QA. Comparable declines are consistently observed across the remaining models as well. These results collectively underscore that existing models still struggle with the complex multi-hop reasoning required to navigate counterfactuals and distractors.

\subsection{Ablation Study}
To better analyze the distinct challenges posed by CRiT-QA, we conduct an ablation study, with results presented in Table~\ref{tab:ablation}. We evaluate model performance by systematically decomposing the context into four settings:
\begin{itemize}
    \item \textit{Oracle ($Q + P_{fact}$)}: The baseline configuration, where the original question $Q$ is paired with the factual gold paragraphs $P_{fact}$ from the source dataset.
    \item \textit{with Counterfactuals ($P_{cf}$)}: The counterfactual configuration, in which factual paragraphs $P_{fact}$ are replaced with generated counterfactual context $P_{cf}$, isolating the challenge of reasoning against parametric knowledge.
    \item \textit{with Distractors ($P_{cf}+P_{dist}$)}: The full CRiT-QA setting, combining counterfactual context $P_{cf}$ with multi-anchor distractor paragraphs $P_{dist}$. 
    \item \textit{without Context (only $Q$)}: A parametric-only probe to confirm that the answer cannot be recalled from memory.
\end{itemize} The results demonstrate the compounded difficulty introduced by each component of CRiT-QA. 
\paragraph{Impact of Counterfactuals.} Transitioning from the \textit{Oracle} setting (59.08 EM for \texttt{Gemini-2.5-Pro}, 34.96 EM for \texttt{Qwen2.5-7B}) to the counterfactual setting $P_{cf}$ leads to substantial degradation in performance. \texttt{Gemini-2.5-Pro} drops by 8.22 EM points to 50.86, while \texttt{Qwen2.5-7B} drops by 10.19 points to 24.77. This reduction highlights the difficulty of overriding memorized factual knowledge and adhering to counterfactual evidence.

\paragraph{Impact of Distractors.} The addition of distractor paragraphs $P_{dist}$ in the full CRiT-QA configuration ($P_{cf}+P_{dist}$) further exacerbates the performance declines. The EM score of \texttt{Gemini-2.5-Pro} falls by an additional 3.12 points to 47.74, while \texttt{Qwen2.5-7B} drops by 5.47 points to 19.30. These results demonstrate the effectiveness of multi-anchor distractor chains in misleading models and underscore the need for more robust evidence-based reasoning.

\paragraph{Context Dependency.} The \textit{without context} setting confirms dataset integrity. Both models record 0.00 EM (with negligible F1 scores of 1.27 and 1.74, respectively), verifying that correct counterfactual answers cannot be retrieved from parametric memory alone. 

Overall, the ablation study demonstrates that performance degradation on CRiT-QA arises not from a single factor but from the compounding challenge of (i) reasoning under counterfactual conditions that conflict with internal knowledge and (ii) resisting plausible multi-hop distractor paths.

\subsection{Analysis by Reasoning Hop Length}
To further investigate the reasoning limitations of current models, we analyze performance on CRiT-QA by breaking down results according to the reasoning chain length (2-hop, 3-hop, and 4-hop), as shown in Table~\ref{tab:benchmark_results_hop}.

This analysis is particularly important given our multi-anchor distractor methodology. In our pipeline, an $N$-hop question is paired with $N$ distinct distractor reasoning chains, each anchored at a different hop. Consequently, a 4-hop question requires not only a longer inference chain but also introduces a larger number of plausible distractor paragraphs $P_{dist}$ than 2-hop or 3-hop questions. Thus, task difficulty is intentionally designed to increase with reasoning length.

The results in Table~\ref{tab:benchmark_results_hop} empirically confirm this design, showing a consistent negative correlation between the number of hops and model performance. Across all evaluated models, accuracy is lowest on 4-hop questions. As reasoning chains lengthen and the distractors accumulate, models become increasingly prone to failure.

Even the most advanced models are not immune to this degradation. For instance, the F1 score of \texttt{GPT-4o} declines from 52.78 (2-hop) to 47.87 (3-hop) and 36.53 (4-hop). \texttt{Claude Sonnet 4.5} shows a similar trend, with its F1 score dropping from 52.82 (2-hop) to 36.07 (4-hop). \texttt{Gemini-2.5-Pro}, while maintaining relatively stable performance between 2-hop (44.89 EM) and 3-hop (45.13 EM) tasks, still exhibits a notable decrease on 4-hop questions (40.74 EM). The decline is even sharper among open-source models. The F1 score of \texttt{LLaMA-3-8B} falls from 28.91 (2-hop) to 10.18 (4-hop), and the EM of \texttt{Qwen2.5-7B} is nearly halved, dropping from 22.80 (2-hop) to 12.84 (4-hop).

These results highlight that models are not only challenged by longer inference chains but also increasingly overwhelmed by the growing density of distractor traps. The cognitive load of simultaneously validating the correct multi-hop path while rejecting multiple plausible, type-consistent distractors represents a key failure point of current LLMs.

\section{Conclusion}

In this paper, we address critical shortcomings in existing multi-hop QA evaluations, where high performance often conceals a reliance on parametric knowledge and dataset-specific shortcuts. We introduce CRiT-QA (Counterfactual Reasoning with Traps), a dataset designed to rigorously evaluate genuine, evidence-driven reasoning by neutralizing memorized knowledge with counterfactual entities and suppressing shallow heuristics through multi-anchor distractor traps. Our experiments empirically validate the effectiveness of CRiT-QA. We demonstrate that current LLMs experience substantial performance degradation on CRiT-QA, standing in sharp contrast to their strong results on standard benchmarks. An ablation study confirms that this degradation arises from the dual challenges of counterfactual adherence and distractor filtering, and that difficulty compounds with increased reasoning depth and distractor density. These findings underscore the persistent gap between surface-level success and genuine multi-hop reasoning. CRiT-QA thus serves as a robust diagnostic tool for exposing these weaknesses and provides a foundation for developing and evaluating next-generation, evidence-grounded LLMs.

\section{Limitations \& Future Directions}
While CRiT-QA provides a robust framework for evaluating multi-hop reasoning, we emphasize that its primary objective is not to establish a static benchmark but to diagnose the surface-level heuristics and memorized parametric knowledge current models rely on. With this diagnostic objective, we identify the following limitations and directions for future research: (i) \textit{Dependency on a Single Base Dataset.} CRiT-QA is currently built upon the MuSiQue dataset, inheriting its specific reasoning structures. However, its fully automated construction pipeline can be readily extended to additional multi-hop benchmarks to encompass a broader range of domains. This automation also enables periodic re-execution to generate new counterfactual entities, effectively mitigating the risk of data contamination. (ii) \textit{Potential Generator Artifacts.} The pipeline relies on an LLM to synthesize both counterfactual contexts and distractor paragraphs, introducing the risk of stylistic artifacts. Models might exploit these as unintended shortcuts instead of performing faithful reasoning. Future work could explore more rigorous artifact-filtering mechanisms to further isolate the reasoning capabilities of evaluated models.

\section{Acknowledgments}

This work was supported by the Institute of Information \& Communications Technology Planning \& Evaluation (IITP) grant funded by the Korea government (MSIT) [RS-2021-II211341, Artificial Intelligence Graduate School Program (Chung-Ang
University)] and by the National Research Foundation of Korea (NRF) grant funded by the Korea
government (MSIT) (RS-2025-00556246).

\section{Bibliographical References}\label{sec:reference}

\bibliographystyle{lrec2026-natbib}
\bibliography{lrec2026-example}

\end{document}